\documentclass[letterpaper, 10 pt, conference]{ieeeconf}  % Comment this line out if you need a4paper

\IEEEoverridecommandlockouts                              % This command is only needed if 
\usepackage{graphics} % for pdf, bitmapped graphics files
\usepackage{epsfig} % for postscript graphics files
\usepackage{amsmath} % assumes amsmath package installed
\usepackage{amssymb}  % assumes amsmath package installed
\usepackage[table]{xcolor}
\usepackage{import}
\usepackage{comment}

\graphicspath{{./pictures/}}

\newcommand{\xFP}{$\mathbf{x}_{FP}$~}
\newcommand{\dxFP}{$\dot{\mathbf{x}}_{FP}$~}

\title{\LARGE \bf
Gaze Stabilization for Humanoid Robots: a Comprehensive Framework
}

\author{Alessandro Roncone$^{1}$, Ugo Pattacini$^{1}$, Giorgio Metta$^{1}$ and Lorenzo Natale$^{1}$% <-this % stops a space
\thanks{*This work was supported by the European Project KoroiBot (FP7-ICT-611909).}% <-this % stops a space
\thanks{$^{1}$A. Roncone, U. Pattacini, G. Metta and L. Natale are with iCub Facility, Istituto Italiano di Tecnologia, Via Morego 30, 16163 Genova, Italy. {\tt\small \{alessandro.roncone, ugo.pattacini, giorgio.metta, lorenzo.natale\}@iit.it}}%
}

\begin{document}

\maketitle
\thispagestyle{empty}
\pagestyle{empty}

%%%%%%%%%%%%%%%%%%%%%%%%%%%%%%%%%%%%%%%%%%%%%%%%%%%%%%%%%%%%%%%%%%%%%%%%%%%%%%%%
\begin{abstract}
Gaze stabilization is an important requisite for humanoid robots. Previous work on this topic has focused on the integration of inertial and visual information. Little attention has been given to a third component, which is the knowledge that the robot has about its own movement. 
In this work we propose a comprehensive framework for gaze stabilization in a humanoid robot. We focus on the problem of compensating for disturbances induced in the cameras due to self-generated movements of the robot. In this work we employ two separate signals for stabilization: (1) an anticipatory term obtained from the velocity commands sent to the joints while the robot moves autonomously; (2) a feedback term from the on board gyroscope, which compensates unpredicted external  disturbances. We first provide the mathematical formulation to derive the forward and the differential kinematics of the fixation point of the stereo system. We finally test our method on the iCub robot. We show that the stabilization consistently reduces the residual optical flow during the movement of the robot and in presence of external disturbances. We also demonstrate that proper integration of the neck DoF is crucial to achieve correct stabilization.
\end{abstract}

%!TEX root = root.tex
\section{INTRODUCTION}\label{sec:intro}
Efficient gaze stabilization in mammals is fundamental because it reduces image blur elicited by the movement of the body during locomotion. The brain senses external motion through the vestibular system and the generated optical flow and performs compensatory movements with the eyes and the head to maintain stable fixation. The effect of the absence of stabilization can be easily measured by taking a picture or shooting a video while walking or running.

Gaze stabilization is therefore a fundamental capability for a humanoid robot. Conventionally, algorithms and behaviors for visual stabilization have been designed drawing inspiration from biological systems. Due to its relative simplicity the brain circuitries involved are relatively well understood~\cite{carpenter1988}. Broadly speaking compensatory movements are obtained with two main contributions. The vestibulo-ocular reflex (VOR) exploits the information about the head movement coming from the vestibular system. The whole control loop in this case involves a few synapses and it is therefore very fast. The opto-kinetic reflex (OKR) uses on the other hand retinal slip from the eyes to generate compensatory movement and maintain stable fixation. The computation in this case involves more complex computations, it has larger latency and is less efficient. However these contributions perform best at different frequencies and are therefore integrated for efficient stabilization.

Early work on oculomotor control in robotics has focused on replicating various type of eye movements like vergence, smooth-pursuit, saccades~\cite{coombs1992,berthouze1996,capurro1997} and gaze stabilization reflexes obtained using inertial and visual input~\cite{panerai98,shibata2000,panerai2002}.

Computation of the eye velocity command for proper stabilization depends on several parameters: eye-head geometry, relative distance between the fixation point and the head but also non-linearities due to lens distortions and delays in the plant. If the eyes and the head do not rotate around the same axes, the compensation signal must take into account the translational velocity due to parallax. This can be done analytically~\cite{panerai98} or with Feedback Error Learning~\cite{shibata2000,panerai2002}. The advantage of the latter methods is that it can also optimally integrate visual and inertial information and compensate for delays in the plant. 

Only in a few cases the attention has been devoted to the problem of gaze stabilization during legged locomotion~\cite{gay2012, oliveira2009}. In~\cite{gay2012} the authors implement a controller based on an oscillator which is adapted to match the frequency and phase of the optical flow generated by the robot gait, in the assumption that the latter is periodic. In~\cite{oliveira2009} the authors use genetic algorithms to evolve a central pattern generator that optimally reduces head shaking during locomotion of a quadruped. 
Previous work on gaze stabilization has focused on the control of the eyes and has ignored a third source of information useful for gaze stabilization, i.e. the motor signals issued to the robot during walking and generic whole-body movements. This information, however, provides important cues for stabilizing motion due to the robot own movement. With respect to inertial and visual signals this information is predictive in that it allows anticipating and planning compensatory movements in advance. 

In this paper we solve the problem of gaze stabilization by integrating a feedback component coming form the sensory system with a feedforward component derived from the commands issued to the motors. We build upon the gaze controller implemented on the iCub~\cite{UgoPhdThesis} and extend it to stabilize gaze during active movements of the iCub~\cite{MettaiCub2010}. The system uses all 6~DoF of the head and it relies on two sources of information: i) the inertial information read from the IMU placed on the robot's head (\textsl{feedback}) and ii) an equivalent signal computed from the commands issued to the motors of the torso (\textsl{feedforward}). For both cues we compute the resulting perturbation of the fixation point and use the Jacobian of the iCub stereo system to compute the motor command that compensates the perturbation. Retinal slip (i.e. optical flow) is used to measure the performance of the system. We show that the feedforward component allows for better compensation of the robot's own movements and, if properly integrated with inertial cues, may contribute to improve performance in presence of external perturbations.  We also show that the DoF of the neck must be integrated in the control loop to achieve good stabilization performance.

The article is structured as follows. In Section \ref{sec:mm}, the proposed framework is defined. The experimental protocol and the related experiments are presented in Section \ref{sec:exp}, followed by Conclusions and Future Work (Section \ref{sec:concl}).
%!TEX root = root.tex
%%%%%%%%%%%%%%%%%%%%%%%%%%%%%%%%%%%%%%%%%%%%%%%%%%%%%%%%%%%%%%%%%%%%%%%%%%%%%%%%
%\section{INTRODUCTION}
%This is the even better introduction we still have to do.

\section{METHOD}\label{sec:mm}

We define the stabilization problem as the stabilization of the 3D position of the fixation point \xFP of the robot. It is achieved by controlling the cameras to keep the velocity \dxFP  equal to zero. The velocity of the fixation point is 6-dimensional, and is composed of a translational component $v_{FP}$ and a rotational part $\omega_{FP}$.

\begin{figure}
\begin{center}
\includegraphics[width=.9\linewidth]{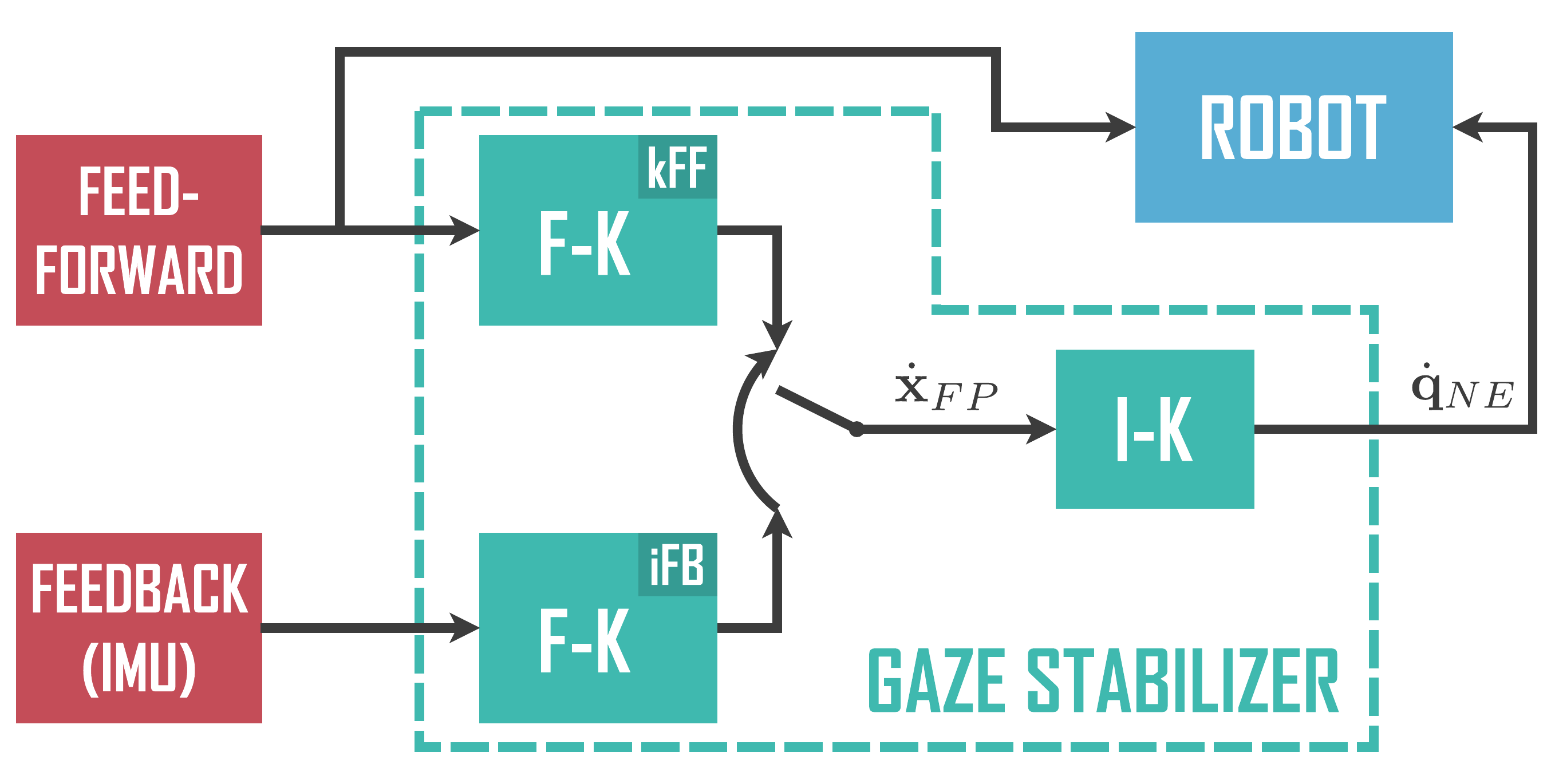}
\caption{Block diagram of the framework presented. The Gaze Stabilizer module (in green) is designed to operate both in presence of a kinematic feedforward (\emph{kFF}) and an inertial feedback (\emph{iFB}). In both cases, it estimates the motion of the fixation point and controls the head joints in order to compensate for that motion.}\label{fig:blockDiagram}
\end{center}
\end{figure}

A diagram of the proposed framework is presented in Fig. \ref{fig:blockDiagram}. As highlighted in Section \ref{sec:intro}, the gaze stabilization module has been designed to operate in two (so far mutually exclusive) scenarios:

\begin{itemize}
\item a \textsl{kinematic feed-forward (kFF)} scenario, in which the robot produces self-generated disturbances due to its own motion; in this case motor commands predict the perturbation of the fixation point and can be used to stabilize the gaze.
\item an \textsl{inertial feed-back (iFB)} scenario, in which perturbations are (partially) estimated by an Inertial Measurement Unit (IMU).
\end{itemize}

As result, the \textsl{Gaze Stabilizer} is realized by the cascade of two main blocks: the first block is used for estimating the 6D motion of the fixation point \dxFP by means of the forward kinematics, while the latter exploits the inverse kinematics of the neck-eye plant in order to compute a suitable set of desired joint velocities $\dot{\mathbf{q}}_{NE}$ able to compensate for that motion. The forward kinematics block represents a scenario-dependent component, meaning that its implementation varies according to the type of input signal (i.e. feed-forward or feedback). Conversely, the inverse kinematics module has a unique realization.

Crucial to this work is the computation of the position of the fixation point and its Jacobian. Section \ref{subs:kinematics} provides a complete formulation of the kinematic problem occurring at the eyes, whereas Section \ref{subs:estimation} and \ref{subs:stabilization} analyze the forward and the inverse kinematics modules composing the Gaze Stabilizer.
\subsection{Forward and Differential Kinematics of the iCub stereo system}\label{subs:kinematics}
\begin{figure}
\begin{center}
\hspace{30pt} % This trick is used to better center-align this figure that is skewed on the left
\includegraphics[width=.7\linewidth]{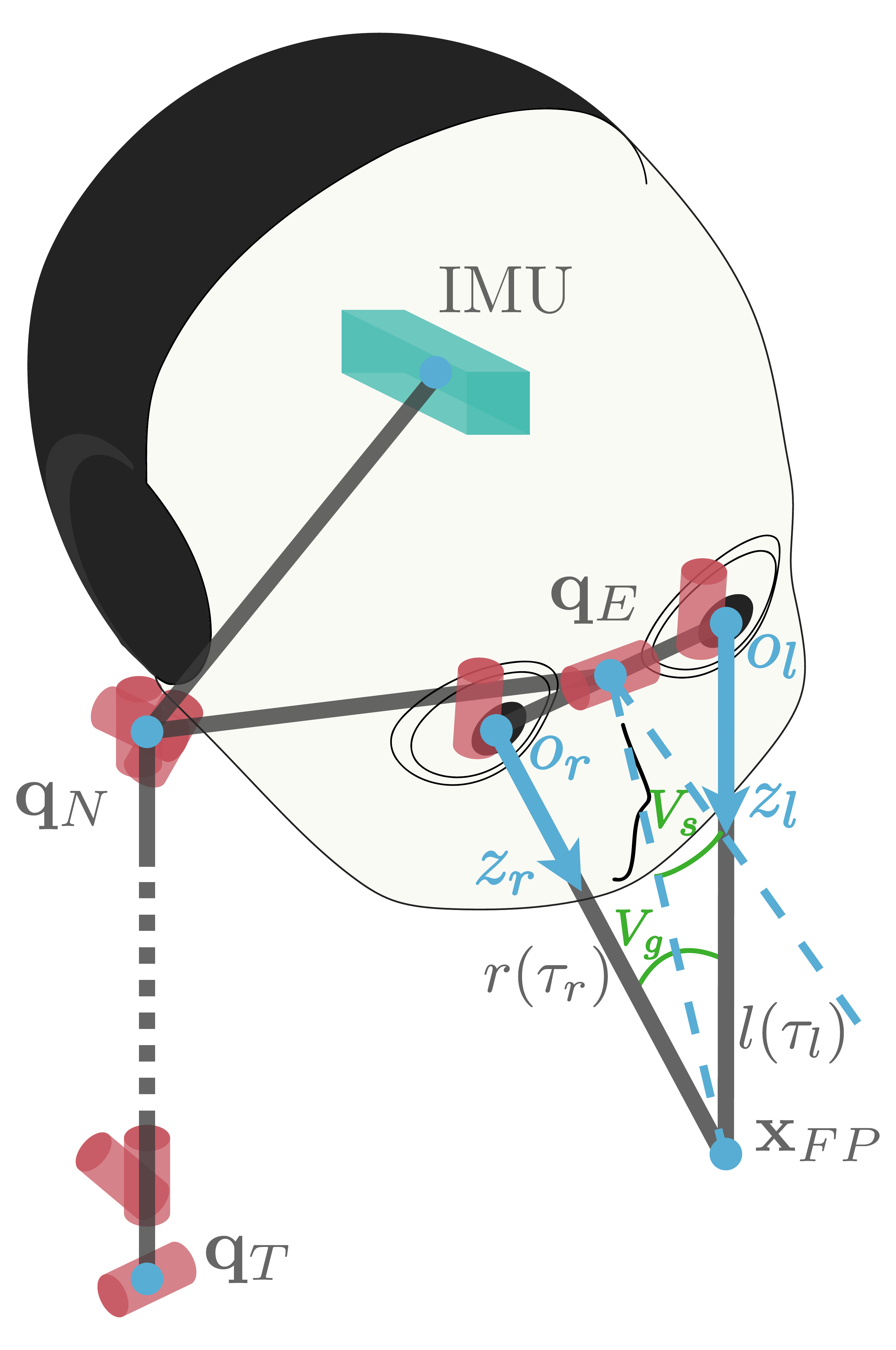}
\vskip -10pt
\caption{Kinematics of the iCub's torso and head. The upper body of the iCub is composed of a 3 DoF torso, a 3 DoF neck and a 3 DoF binocular system, for a total of 9 DoF. Each of these joints, depicted in red, are responsible for the motion of the fixation point. The Inertial Measurement Unit (IMU) is the green rectangle placed in the head; its motion is not affected by the eyes.}\label{fig:kinematics}
\end{center}
\end{figure}

To derive the Jacobian of the fixation point we start from the forward kinematic law of the eyes as illustrated in Fig. \ref{fig:kinematics}. The position of the fixation point \xFP is computed in two steps. The first step computes the position of the frame of reference of the eyes. This uses a representation of the forward kinematics of the iCub head in standard Denavit-Hartenberg notation (the DH parameters of the iCub are reported here: \cite{UgoPhdThesis}). The second step computes \xFP  as the intersection of the two rays joining the cameras optical centers and the projection of the target on the camera planes.
\subsubsection{Forward Kinematics}\label{subsubs:fkfw}
by referring to Figure \ref{fig:kinematics}, the 3D Cartesian position of the fixation point \xFP \ can be intuitively defined as the intersection point of the lines $l(\tau_l)$ and $r(\tau_r)$ that originate from the left and right camera planes passing through the respective optical centers. In a parametric formulation, they are defined as:
\begin{equation}
\begin{aligned}
l(\tau_l): o_l &+ \tau_l \cdot z_l\\
r(\tau_r): o_r &+ \tau_r \cdot z_r
\end{aligned}\label{eq:lr}\ ,
\end{equation}
where $o_l$ and $o_r$ are the centers of the left and right camera planes respectively, and $z_l$ and $z_r$ are the axes perpendicular to these planes, as shown in Figure \ref{fig:kinematics}. To address the more general case of skew lines (i.e. $l(\tau_l)$ and $r(\tau_r)$ might not be coplanar due to mechanical misalignments of image planes), the fixation point \xFP can be defined as the mean point of the shortest segment between $l(\tau_l)$ and $r(\tau_r)$. From Eq. \ref{eq:lr}, it is possible to derive the points $\mathbf{P}_l$ and $\mathbf{P}_r$ that belong to each line and minimize the distance from the other line. They are given by:
\begin{equation}
\begin{aligned}
\tau_l^*= \frac{\left[ z_l - \left( z_l \cdot z_r \right) \cdot z_r \right]\cdot\left[ o_l - o_r \right]}{\left( z_l \cdot z_r \right)^2 -1}  \\
\tau_r^*= \frac{\left[ \left( z_l \cdot z_r \right) \cdot z_l - z_r \right]\cdot\left[ o_l - o_r \right]}{\left( z_l \cdot z_r \right)^2 -1}
\end{aligned}\label{eq:tlrstar}\ .
\end{equation}
Finally, the intersection point \xFP can be found as the mean point between $\mathbf{P}_l$ and $\mathbf{P}_r$:
\begin{equation}
\mathbf{x}_{FP} = \frac{\mathbf{P}_l+\mathbf{P}_r}{2} = \frac{o_l + \tau_l^* \cdot z_l + o_r + \tau_r^* \cdot z_r}{2}\ .
\label{eq:xfp}
\end{equation}
\subsubsection{Differential Kinematics}\label{subsubs:fkdiff}
the position of the fixation point in the Cartesian space depends on the whole body configuration, namely the legs, the torso, the neck and the eyes: $\mathbf{q} = \left[ \mathbf{q}_{L},\ \mathbf{q}_{T},\ \mathbf{q}_{N},\ \mathbf{q}_{E} \right]^T$. It is possible to profitably apply the standard DH notation to the kinematics of all the body parts with the exception of the eyes. On the iCub, indeed, three DoFs (the common tilt $t_c$, the version $v_s$ and the vergence $v_g$) account for four coupled joints actuating the eyes (the tilt and pan for the left and right cameras, i.e. $\left[\mathtt{t}_l,\ \mathtt{p}_l\right]^T$ and $\left[\mathtt{t}_r,\ \mathtt{p}_r\right]^T$ respectively). In particular, $\mathbf{q}_E$ is given by:
\begin{equation}
    \mathbf{q}_{E} =
    \left[
    \begin{aligned}
    &t_c \\
    &v_s \\
    &v_g
    \end{aligned}
    \right]
    =
    \left[
    \begin{aligned}
    \mathtt{t}_l = \mathtt{t}_r \\
    \frac{\mathtt{p}_l+\mathtt{p}_r}{2} \\
    \mathtt{p}_l-\mathtt{p}_r
    \end{aligned}
    \right] \label{eq:qejoints}\ ,
\end{equation}
and this leads to the inverse relations:
\begin{equation}
\mathtt{t}_l = \mathtt{t}_r = t_c, \quad
\mathtt{p}_l = v_s + v_g/2, \quad
\mathtt{p}_r = v_s - v_g/2
\label{eq:qedof}\ .
\end{equation}
For what concerns the motion of the fixation point \dxFP, for the purposes of this work we are only interested in finding the relation between the joints velocities $\dot{\mathbf{q}}_E$ and its translational component $v_{FP}$, as detailed in Section \ref{subs:stabilization}.
Under this assumption, the Jacobian matrix $J_E$ that relates the motion of the fixation point \xFP with the eyes joints $\mathbf{q}_E$ will be reduced to a $3\times3$ matrix.
The standard analytical Jacobian matrix is defined as:
\begin{equation}
    J_E = \frac{\partial \mathbf{x}_{FP}(\mathbf{q}_E)}{\partial \mathbf{q}_E} =
    \left[
    \frac{\partial \mathbf{x}_{FP}}{\partial t_c},\quad \frac{\partial \mathbf{x}_{FP}}{\partial v_s},\quad \frac{\partial \mathbf{x}_{FP}}{\partial v_g}
    \right]\ .
\end{equation}
Using the chain rule, and Equations \ref{eq:xfp} and \ref{eq:qedof}, leads to:
\begin{subequations}
\begin{small}\begin{align}
\frac{\partial \mathbf{x}_{FP}}{\partial t_c} &= \frac{\partial \mathbf{x}_{FP}}{\partial t_c} = \frac{1}{2}\left( \frac{\partial \mathbf{P}_l}{\partial t_c} + \frac{\partial \mathbf{P}_r}{\partial t_c}\right) \label{eq:partialT} \\
\frac{\partial \mathbf{x}_{FP}}{\partial v_s} &=  \frac{\partial\mathbf{x}_{FP}}{\partial\mathtt{p}_l} + \frac{\partial\mathbf{x}_{FP}}{\partial\mathtt{p}_r} = \nonumber \\
 &= \frac{1}{2}\left( \frac{\partial\mathbf{P}_l}{\partial\mathtt{p}_l} + \frac{\partial\mathbf{P}_l}{\partial\mathtt{p}_r} + \frac{\partial\mathbf{P}_r}{\partial\mathtt{p}_l} + \frac{\partial\mathbf{P}_r}{\partial\mathtt{p}_r} \right) \label{eq:partialVs} \\
\frac{\partial \mathbf{x}_{FP}}{\partial v_g} &=  \frac{1}{2}\cdot \left(\frac{\partial\mathbf{x}_{FP}}{\partial\mathtt{p}_l} - \frac{\partial\mathbf{x}_{FP}}{\partial\mathtt{p}_r} \right) = \nonumber \\ 
&= \frac{1}{4}\cdot \left(\frac{\partial \mathbf{P}_l}{\partial\mathtt{p}_l} - \frac{\partial \mathbf{P}_l}{\partial\mathtt{p}_r} + \frac{\partial \mathbf{P}_r}{\partial\mathtt{p}_l} - \frac{\partial \mathbf{P}_r}{\partial\mathtt{p}_r} \right)\label{eq:partialVg}\ .
\end{align}\end{small}
\end{subequations}

The computation of the quantities presented in Equations \ref{eq:partialT}, \ref{eq:partialVs} and \ref{eq:partialVg} depends from Equations \ref{eq:xfp} and \ref{eq:tlrstar}. For simplicity we derive only the first factor of Eq. \ref{eq:partialT}; the derivation of the other components has been omitted for brevity but can be derived similarly. $\partial \mathbf{P}_l/\partial t_c$ is given by:
\begin{equation}
\frac{\partial \mathbf{P}_l}{\partial t_c} = \frac{\partial \left( o_l + \tau_l^* \cdot z_l \right)}{\partial t_c} = \frac{\partial o_l}{\partial t_c} + \frac{\partial \tau_l^* }{\partial t_c}\cdot z_l + \tau_l^* \cdot \frac{\partial z_l}{\partial t_c}\ .
\end{equation}
$\partial o_l / \partial t_c$ and $\partial z_l / \partial t_c$ represent, respectively, the geometric Jacobian of the left eye and the analytical Jacobian of the z-axis of the left eye with respect to the tilt; they are described in Equation \ref{eq:ozlr}. The second derivative is instead more complex. Let us define:
\begin{equation}
\begin{aligned}
\xi_0 &= z_l \cdot z_r \\ 
\xi_1 &= \left[ z_l - \left( z_l \cdot z_r \right) \cdot z_r \right] = \left[ z_l - \xi_0 \cdot z_r \right] \\ 
\xi_2 &= \left[ o_l - o_r \right] \\
\xi_3 &= \left( z_l \cdot z_r \right)^2 -1 = \left( \xi_0 \right)^2 -1\quad ;
\end{aligned}\label{eq:q0123}
\end{equation}
thus, $\partial \tau_l^* / \partial t_c$ becomes:
\begin{equation}
\begin{aligned}
\frac{\partial \tau_l^* }{\partial t_c} &= \frac{\partial}{\partial t_c} \left( \frac{\xi_1 \cdot \xi_2}{\xi_3} \right) = \\ 
&= \frac{\xi_2 \xi_3 \partial \xi_1 / \partial t_c + \xi_1 \xi_3 \partial \xi_2 / \partial t_c - \xi_1 \xi_2 \partial \xi_3 / \partial t_c}{\xi_3^2}\ .
\end{aligned}
\end{equation}
Finally, $\partial \xi_1 / \partial t_c$, $\partial \xi_2 / \partial t_c$, and $\partial \xi_3 / \partial t_c$ can be derived from Equation \ref{eq:q0123} and are compositions of:
\begin{equation}
\begin{aligned}
\frac{\partial o_l }{\partial t_c} &= J^G_l(t_c) \qquad \ 
\frac{\partial z_l }{\partial t_c} &= J^A_l(t_c) \\ 
\frac{\partial o_r }{\partial t_c} &= J^G_r(t_c) \qquad \ 
\frac{\partial z_r }{\partial t_c} &= J^A_r(t_c)    
\end{aligned}\label{eq:ozlr}\quad ,
\end{equation}
where $J^G_l(t_c)$ and $J^G_r(t_c)$ are the geometric Jacobians of the left and right camera optical centers with respect to the common tilt, whereas $J^A_l(t_c)$ and $J^A_r(t_c)$ are the analytical Jacobians of the left and right z-axis with respect to the tilt. Both $J^A$ and $J^G$ can be retrieved with resort to the standard kinematics libraries as in \cite{UgoPhdThesis}.
\subsection{Estimating the motion of the fixation point}\label{subs:estimation}
As discussed in Sections \ref{sec:intro} and \ref{sec:mm}, in this work we exploited the gaze stabilization in two different scenarios, described in the following Subsections.

\subsubsection{Kinematic Feedforward}\label{subsubs:kinfwd}
in the first scenario the robot moves autonomously its body and we estimate the motion of the fixation point with resort to the kinematic model of the robot \cite{UgoPhdThesis}. Under these assumptions, the task is completely defined: given the joints velocities that the robot is actuating at the motors, the fixation point is moving according to the Jacobian of the kinematic chain under consideration. As an example, let us assume that the robot has fixed hips (i.e. no movement at the lower limbs) and is exerting a given set of velocities at the torso ($\dot{\mathbf{q}}_T$), neck ($\dot{\mathbf{q}}_N$) and eyes ($\dot{\mathbf{q}}_E$). At any given instant of time, the motion of the fixation point is given by:
\begin{equation}
    \dot{\mathbf{x}}_{FP} =
    \left[
    \begin{aligned}
    v_{FP}\\
    \omega_{FP}
    \end{aligned}
    \right]
     = J_{TNE} \cdot
    \left[
    \begin{aligned}
    \dot{\mathbf{q}}_T \\
    \dot{\mathbf{q}}_N \\
    \dot{\mathbf{q}}_E
    \end{aligned}
    \right] \label{eq:1}\quad ,
\end{equation}
where $J_{TNE}$ is the $6\times9$ Jacobian of the forward kinematics map relative to the torso, the neck and the eyes.

\subsubsection{IMU Feedback}\label{subsubs:imu}
in the second application, we exploited the measurements provided by the IMU device to estimate the motion occurring at the head. The iCub head is currently equipped with the MTx sensor from Xsens \cite{xsense}, whose location with respect to the robot kinematic is known \cite{UgoPhdThesis}. Among the various sensing elements available from such device, the one of interest here is the gyroscope, able to estimate the 3D rotational velocity $\omega_{IMU}$ of the sensor at any given instant of time. From this measurement, it is possible to derive the 6D velocity of the fixation point \dxFP :
\begin{subequations}
\begin{align} 
v_{FP} &= \omega_{IMU} \times \mathbf{r},\qquad \mathbf{r} = \mathbf{x}_{FP} - \mathbf{x}_{IMU}    \label{eq:2v}  \\ 
\omega_{FP} &= \omega_{IMU}    \ , \label{eq:2o}
\end{align}
\end{subequations}
where $v_{FP}$ is the 3D translational velocity of the fixation point, $\omega_{FP}$ is its 3D rotational velocity, and $\mathbf{r}$ is the lever arm between the position of the fixation point \xFP and the position of the inertial sensor $\mathbf{x}_{IMU}$. It is worth noticing that this is a sub-optimal case: since the inertial sensor measures only a 3D rotational velocity (i.e. $\omega_{IMU}$), we do not have access to the 3D translational component $v_{IMU}$. In this scenario we can only compensate for the the rotational velocity as it is measured by the sensor (Eq. \ref{eq:2o}) and its effect on the translational component (Eq. \ref{eq:2v}). 

\subsection{Gaze stabilization from the estimation of the fixation point motion}\label{subs:stabilization}

In the previous sections we illustrated how the feedforward and feedback terms produce an estimation of the velocity of the fixation point $\dot{\mathbf{x}}_{FP} = \left[ v_{FP},\ \omega_{FP} \right]^T$. Using the inverse kinematics we derive the compensatory motor commands for the head (see Figure~\ref{fig:blockDiagram}):
%. Assuming that the motion of the fixation point $\dot{\mathbf{x}}_{FP} = \left[ v_{FP},\ \omega_{FP} \right]^T$ is given, the control problem is solved by the exact inverse kinematics law:
\begin{equation}
    \left[              
    \begin{aligned}
    \dot{\mathbf{q}}_N \\
    \dot{\mathbf{q}}_E
    \end{aligned}
    \right]
    = - J^{\#}_{NE} \cdot
    \left[
    \begin{aligned}
    v_{FP}\\
    \omega_{FP}
    \end{aligned}
    \right] \label{eq:iKK}\ ,
\end{equation}
where $J^{\#}_{NE}$ is the $6\times6$ pseudo-inverse of the Jacobian of the forward kinematics map relative to the neck and the eyes, and $\dot{\mathbf{q}}_N$, $\dot{\mathbf{q}}_E$ are the desired joint velocities at the neck and eyes respectively.

In this work, we chose to decouple the inverse kinematics problem into two sub-problems: instead of using the full 6-DoF chain of the neck and the eyes to stabilize the 6-DoF motion of the fixation point, we designed the controller such that the neck compensates the rotational component $\omega_{FP}$, whilst the eyes have to counterbalance the translational part $v_{FP}$. The reason is twofold: 1) the neck and the eyes exhibit two different dynamics, the eyes being faster than the neck joints; 2) it is not physically possible for the neck joints alone to stabilize the translational motion $v_{FP}$ and, similarly, the eyes chain can not compensate for the roll of the fixation point by mechanical design. Hence, Equation \ref{eq:iKK} has been split into:
\begin{equation}
    \left\{
    \begin{aligned}
    \dot{\mathbf{q}}_N &= - J^{\#}_{N} \cdot \omega_{FP} \\
    \dot{\mathbf{q}}_E &= - J^{\#}_{E} \cdot v_{FP}
    \end{aligned}
    \right.\ ,
\end{equation}
with $J^{\#}_{N}$ and $J^{\#}_{E}$ being the two independent $3\times3$ pseudo-inverse matrices of the neck and the eyes respectively. The computed joint velocities $\dot{\mathbf{q}}_N$, $\dot{\mathbf{q}}_E$ are then used as reference signals by the joint-level PID controllers.

This decoupling is beneficial for the stability of the system and it does not affect the final performance. The neck and the eyes are controlled to compensate two different components of the motion of the fixation point but cooperate to achieve the task. The rotational motion that is not compensated by the neck in fact produces translational velocities of the fixation point that are compensated by the eyes.
%!TEX root = root.tex
\section{EXPERIMENTAL RESULTS}\label{sec:exp}
\noindent To validate our work we set up two experiments:

\begin{itemize}
\item \textit{Exp. A: compensation of self-generated motion}: we issue a predefined sequence at the yaw, pitch, and roll of the torso and test both the \emph{iKK} and the \emph{iFB} conditions to proved a repeatable comparison between the two.
\item \textit{Exp. B: compensation in presence of an external perturbation}: the motion of the fixation point is caused by the experimenter who physically moves the torso of the robot. In this case there is no feedforward signal available, and the robot uses only the \emph{iFB} signal.
\end{itemize}

\noindent For each experiment, two different sessions have been conducted: in the first session the robot stabilizes the gaze only with the eyes, while in the second session it uses both the neck and the eyes.
In both the scenarios, a session without compensation has been performed and used as a baseline for comparison. It is worth noticing that Experiment A is obviously a more controlled scenario, and for this reason we have used it to obtain a quantitative analysis. In Experiment B instead the disturbances are generated manually, and, as such, it provides only a qualitative assessment of the performance of the \emph{iFB} modality.

For validation we use the dense optical flow measured from the cameras. This can be used as an external, unbiased measure because as explained in Section~\ref{sec:intro} it is not used in the stabilization loop.
We used the OpenCV~\cite{opencv_library} implementation of the dense optical flow algorithm proposed by Farneback~\cite{farneback2003}. Given an input image at time $t$, the method finds the 2D optical flow vector $\mathtt{of}(u,v)$ for each pixel in the image.
%
%\begin{align}
%p^{t-1}(u,v) &\sim p^{t}\left(u+\mathtt{of}^t(u,v)[0],v+\mathtt{of}^t(u,v)[1]\right) \label{eq:of}\\
%|| \mathtt{of}^t(u,v) || &= \sqrt{\mathtt{of}^t(u,v)[0]^2+\mathtt{of}^t(u,v)[1]^2} \label{eq:ofmodule}
%\end{align}
%where $\mathtt{of}(u,v)$ is the two-dimensional optical flow for that pixel. Its module, computed in Equation \ref{eq:ofmodule}, can be used to extrapolate a global estimate of the optical flow, by computing an average value over the whole image:
We derive a measure of performance by averaging the norm of the motion vectors $\mathtt{of}(u,v)$ in the whole image, i.e.:

\begin{equation}
\mathtt{optFl}(t) = \frac{1}{W-40\times~H-40} \sum_{u=20}^{W-20} \sum_{v=20}^{H-20} || \mathtt{of}^t(u,v) || \label{eq:optFlow}\ ,
\end{equation}
in which we remove from the computation the optical flow vectors of the peripheral region of the image. The reason for this is to compute a performance index that is more appropriate for the task, given that the gaze stabilization is computed for the fixation point (in this work $W=320$, $H=240$).
%
%
%and $w$ and $h$ the size of  (respectively, $320$ and $240$ pixels). As it is possible to evince from Equation \ref{eq:optFlow}, for the purposes of this work the optical %flow has been computed only by selecting the central part of the camera plane. This has been done in order to provide an estimation that is as appropriate to the task as possible, since it reduces the computation only in the area in which the gaze stabilization should have an effect, i.e. the foveal region.

\begin{figure}
\begin{center}
\includegraphics[width=.95\linewidth]{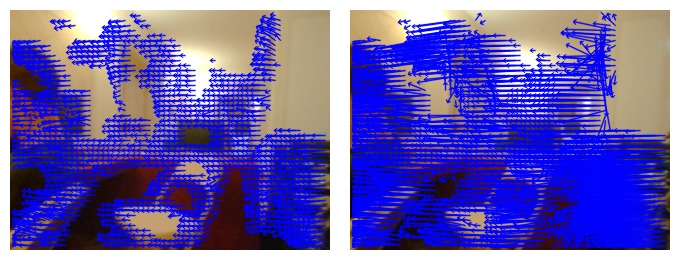}
\vskip -10pt
\caption{Optical flow computed from two subsequent image frames from the left camera, baseline experiment (no compensation). Blue 2D arrows represent the optical flow vector $\mathtt{of}^t(u,v)$ at each pixel. For clarity optical flow vectors are reported only for a subset of the pixels (one pixel every five).}\label{fig:optFlowA}
\vskip 10pt
\includegraphics[width=.95\linewidth]{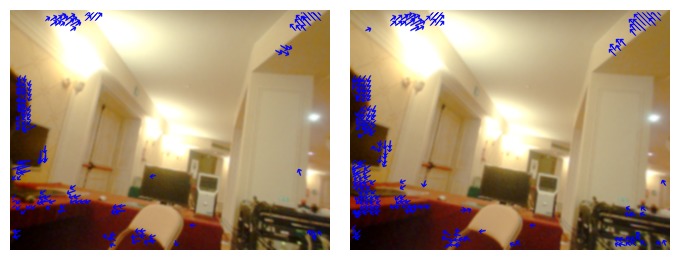}
\vskip -10pt
\caption{Optical flow computed fro two subsequent image frames from the left camera, \emph{iFB} experiment (compensation using inertial feedback). Blue 2D arrows represent the optical flow vector $\mathtt{of}^t(u,v)$ at each pixel. For clarity optical flow vectors are reported only for a subset of the pixels (one pixel every five).}\label{fig:optFlowB}
\end{center}
\end{figure}

The optical flow computed during an experimental session is shown in Figure \ref{fig:optFlowA} and \ref{fig:optFlowB} for two consecutive frames in the baseline experiment (no compensation) and the iFb experiment (stabilization with inertial feedback) respectively. This qualitative evaluation shows that the stabilization effectively reduces the motion in the images. In the following Sections we provide a quantitative evaluation of our framework.

\begin{figure}[t!]
\begin{center}
\includegraphics[width=.95\linewidth]{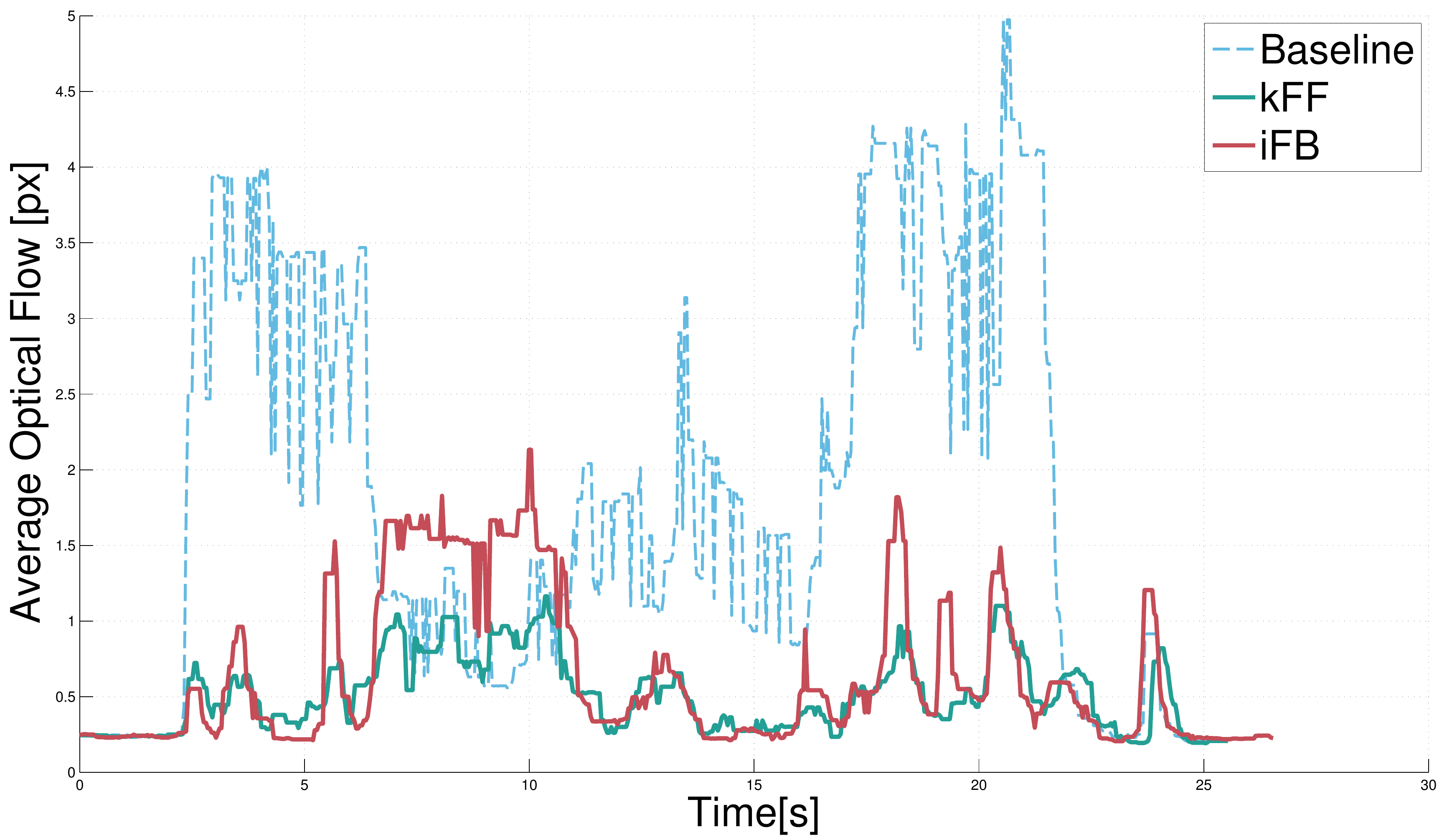}
\vskip -4pt
\caption{Average Optical Flow during Experiment A. In this case only the eyes are controlled. The baseline session is the dashed blue line, while the \emph{kFF} and \emph{iFB} conditions are green and the red lines respectively.}\label{fig:ExpA_Eyes}
\end{center}
\end{figure}

\begin{figure}[t!]
\begin{center}
\includegraphics[width=.95\linewidth]{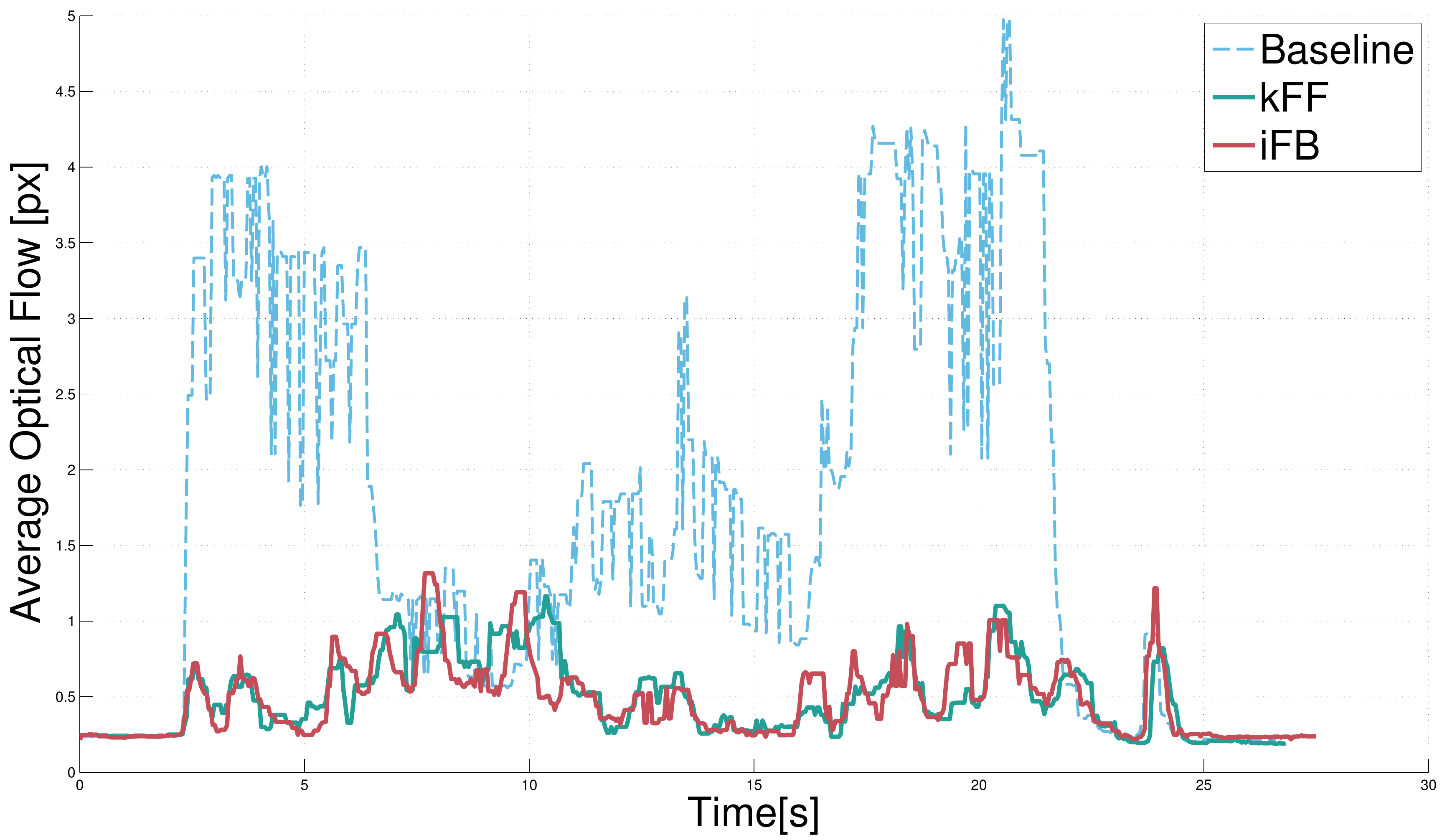}
\vskip -4pt
\caption{Average optical flow during Experiment A. In this case stabilization uses all 6~Dof of the head. The baseline behavior is the dashed blue line, while the \emph{kFF} and \emph{iFB} conditions are green and red lines respectively.}\label{fig:ExpA_NeckEyes}
\end{center}
\end{figure}

\subsection{Compensation in presence of predefined torso movements}
In experiment A we generate a set of predefined movements with the torso. We then compare the \emph{kFF} and the \emph{iFB} conditions with respect to the baseline. In all three cases we use the same sequence of velocity commands to the three torso joints (yaw, pitch and roll). Joints have been controlled with a velocity commands of $20\ \mathtt{deg}/\mathtt{s}$) first independently and then simultaneously. 
As discussed in Section \ref{sec:exp}, the controller has been tested in two cases: using only the 3~DoF of the eyes, and using all 6~DoF composed by the neck and the eyes. Figures \ref{fig:ExpA_Eyes} and \ref{fig:ExpA_NeckEyes} report the average optical flow $\mathtt{optFl}(t)$ in the two conditions respectively.

\begin{figure}
\begin{center}
\includegraphics[width=.65\linewidth]{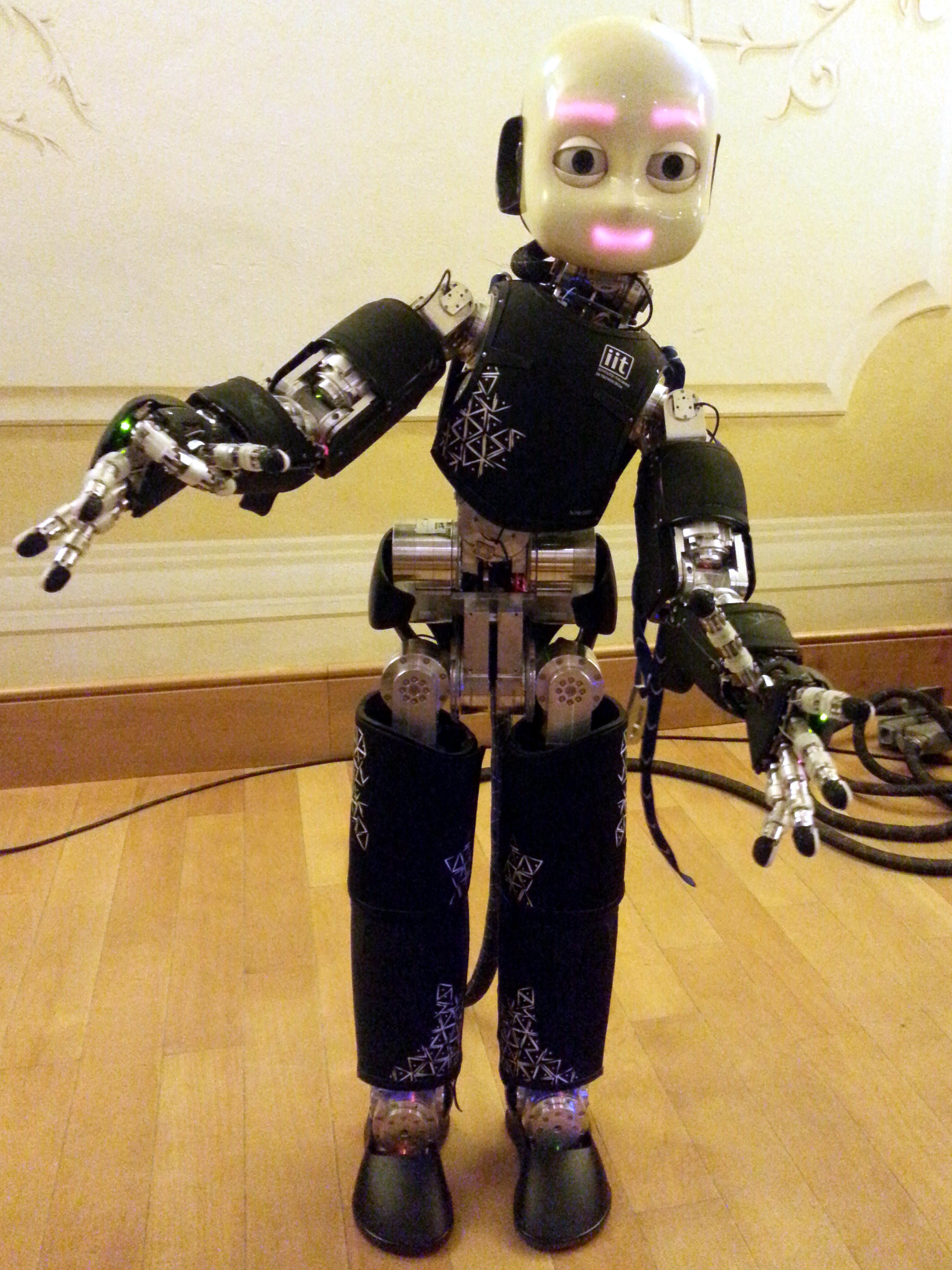}
\vskip -4pt
\caption{The iCub compensating for the roll movement at the torso (Exp A, \emph{kFF} scenario). In this particular occurrence, the stabilization is possible only with respect to the rotational component $\omega^{roll}_{FP}$, since it is not physically feasible for the eyes to compensate such a movement.}\label{fig:iCubRoll}
\end{center}
\end{figure}

The two plots show the improvement of the stabilization with respect to the baseline ($68.1\%$ on average). As expected, the system performed better in the \emph{kFF}  condition than in the the \emph{iFB} case ($23.1\%$ on average): this is because in the former case the system uses a feedforward command that anticipates and better compensates for the disturbances at the fixation point \xFP. Furthermore, a comparison between Figure~\ref{fig:ExpA_Eyes} and Figure~\ref{fig:ExpA_NeckEyes} confirms that by exploiting all 6~DoFs in the head, the performance of the system improves by $24.4\%$ on average. This occurs in particular when, during the sequence, the robot performs a large movement along the roll with the torso (roughly between $t=6s$ and $t=10s$, see also Figure \ref{fig:iCubRoll}). In this situation the optical flow in both the \emph{kFF} and the \emph{iFB} conditions has a peak because the disturbance cannot be compensated with the eyes. Indeed in this case the stabilization fails completely and actually produces unwanted motion (optical flow is higher than the baseline). Notice by comparison with Figure~\ref{fig:ExpA_NeckEyes} that stabilization is more effective when the robot can exploit the additional DoFs of the neck.
%
%of Figure \ref{fig:ExpA_Eyes} have a peak, namely the roll movement (shown in Figure \ref{fig:iCubRoll}); during such a movement, indeed, they eyes alone can not %compensate for the motion at the fixation point at all, and they are actually performing worse than if there were no stabilization at all.

\subsection{Compensation of unknown disturbances}

In experiment B the motors of the joints have been deactivated to allow a human operator to produce disturbances by manually shaking the torso. This is by design a non-repeatable experiment, but it can act as a confirmation of the performances of the \emph{iFB}. As for Experiment A the improvement of the stabilization with respect to the baseline are remarkable ($78\%$ on average), with an improvement of $21.4\%$ when the robot uses all 6~DoF of the head.

\begin{figure}
\begin{center}
\includegraphics[width=.90\linewidth]{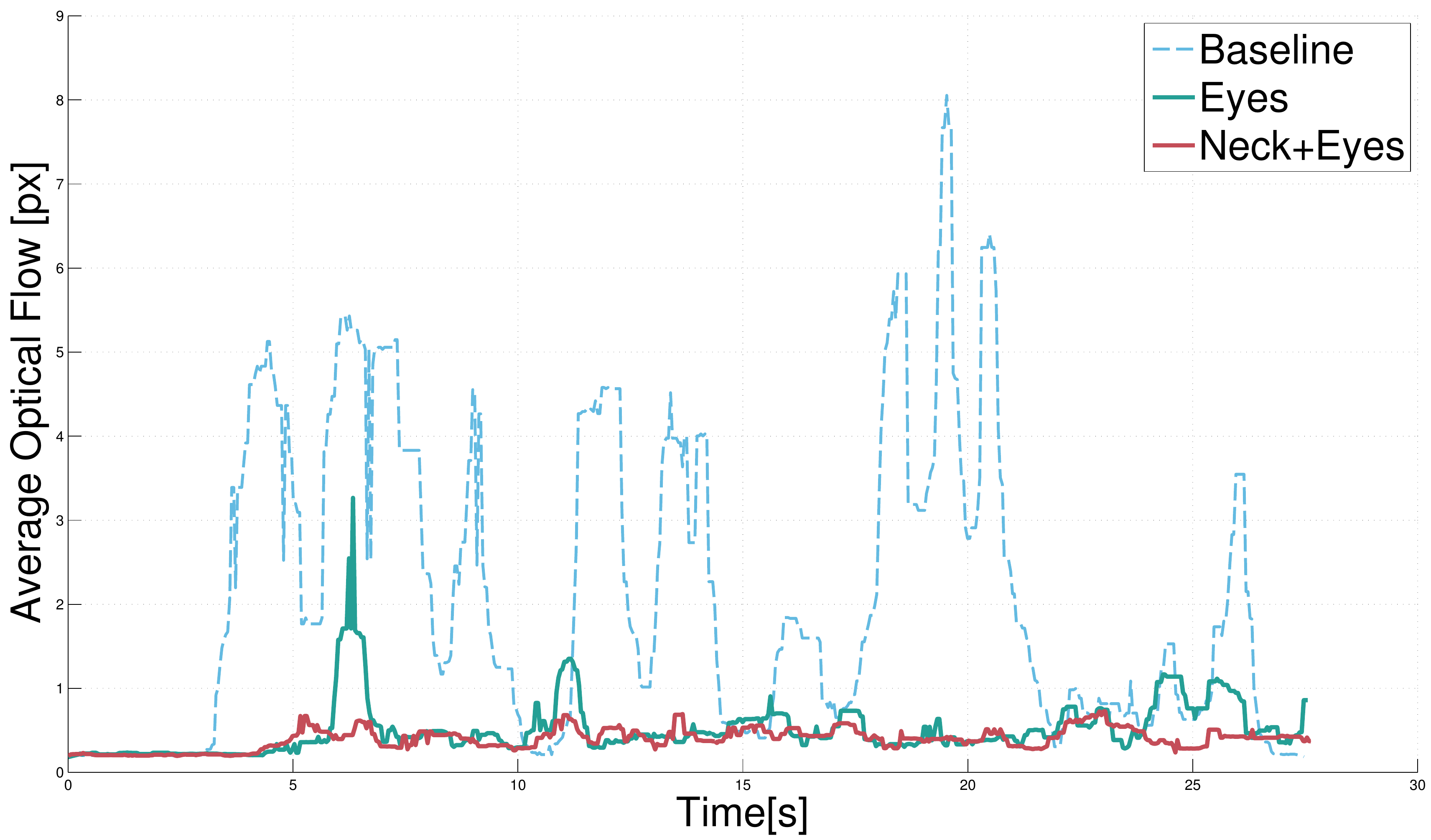}
\vskip -4pt
\caption{Average optical flow during Experiment B. The blue dashed line represents the baseline. Green line is the optical flow when the stabilization uses only the eyes while green line is the optical flow when the stabilization uses all 6~DoF of the head.}\label{fig:ExpB}
\end{center}
\end{figure}

\section{CONCLUSIONS AND FUTURE WORK}\label{sec:concl}

In this paper we described a framework for gaze stabilization of a humanoid robot. With respect to previous work we focus on the use of feedforward commands derived from the knowledge of the motor commands issued to the robot to improve stabilization when perturbations are generated by the robot own movements (e.g. locomotion or generic whole-body motion). To compensate for external perturbations we also include a feedback component provided by the inertial unit mounted on the head of the robot. Our experiments demonstrate that the feedforward component is effective for stabilization when perturbations are due to the robot's own movement. We also demonstrate that proper integration of the DoFs of the neck in the control loop is crucial to achieve good stabilization.
%
%gaze stabilization technique that does not necessarily rely on the typical sources of information used for compensating the motion occurring at the head. Instead of relying solely on a feedback, we suggested a new source of information, namely the kinematic feedforward (\emph{kFF}) computed from the motor commands issued to the robot, to be used side by side with the inertial feedback (\emph{iFB}) coming from the IMU readings. 
%
%We defined the gaze stabilization as the minimization of the motion occurring at the fixation point \xFP, and we presented the direct and differential kinematics relative to the non standard kinematic chain occurring at the iCub head (and, generally, at any binocular head).
%
%Furthermore, we demonstrated that the \emph{kFF} scenario performs globally better than the \emph{iFB}, although it is not as general purpose as the \emph{iFB}. It is worth noticing, though, %that both the approaches significantly improved the baseline behavior, with a reduction of the Average Optical Flow on the image of up to $82\%$.
%

In the experiments reported in this paper the robot compensated disturbances induced only by the motion of the upper body and we did not integrate the feedback and feedforward components. In addition optical flow was not used for the stabilization but only as a performance measure. This is therefore only a first step in the implementation of a full gaze stabilization system for a humanoid robot. As part of our future work we will investigate how to optimally integrate feedforward information with feedback coming from the inertial system and optical flow from the cameras.  Furthermore, a natural extension of this framework is to integrate the information from the whole body of the iCub, including feedforward commands for all motors, feedback from the inertial units, torque sensors at the arms and legs as well as the tactile feedback from the skin.

%
%
%
%a variation of the proposed framework that will effectively take advantage of the higher degree of information it is subject to: the fusion of the signal coming from the \emph{kFF} with the \emph{iFB} will pave the way for a more robust, comprehensive behavior able to take advantage of both the modalities presented in this work. Under these assumptions, the optical flow can be used as a third source of feedback as well.

% \addtolength{\textheight}{-12cm}  % This command serves to balance the column lengths
                                  % on the last page of the document manually. It shortens
                                  % the textheight of the last page by a suitable amount.
                                  % This command does not take effect until the next page
                                  % so it should come on the page before the last. Make
                                  % sure that you do not shorten the textheight too much.

%\section*{ACKNOWLEDGMENT}

\bibliographystyle{IEEEtran}
\bibliography{IEEEabrv,bibliography}

\begin{thebibliography}{10}
\providecommand{\url}[1]{#1}
\csname url@rmstyle\endcsname
\providecommand{\newblock}{\relax}
\providecommand{\bibinfo}[2]{#2}
\providecommand\BIBentrySTDinterwordspacing{\spaceskip=0pt\relax}
\providecommand\BIBentryALTinterwordstretchfactor{4}
\providecommand\BIBentryALTinterwordspacing{\spaceskip=\fontdimen2\font plus
\BIBentryALTinterwordstretchfactor\fontdimen3\font minus
  \fontdimen4\font\relax}
\providecommand\BIBforeignlanguage[2]{{%
\expandafter\ifx\csname l@#1\endcsname\relax
\typeout{** WARNING: IEEEtran.bst: No hyphenation pattern has been}%
\typeout{** loaded for the language `#1'. Using the pattern for}%
\typeout{** the default language instead.}%
\else
\language=\csname l@#1\endcsname
\fi
#2}}

\bibitem{carpenter1988}
\BIBentryALTinterwordspacing
{Roger H.S. Carpenter}, \emph{Movements of the eyes}, ser. Medical.\hskip 1em
  plus 0.5em minus 0.4em\relax London, UK: Pion, 1988. [Online]. Available:
  \url{http://www.ccsds.org/documents/pdf/CCSDS-101.0-B-4.pdf}
\BIBentrySTDinterwordspacing

\bibitem{coombs1992}
D.~Coombs and C.~Brown, ``Real-time smooth pursuit tracking for a moving
  binocular robot,'' in \emph{Computer Vision and Pattern Recognition, 1992.
  Proceedings CVPR '92., 1992 IEEE Computer Society Conference on}, Jun 1992,
  pp. 23--28.

\bibitem{berthouze1996}
L.~Berthouze, S.~Rougeaux, F.~Chavand, and Y.~Kuniyoshi, ``Calibration of a
  foveated wide-angle lens on an active vision head,'' in \emph{Computer Vision
  and Pattern Recognition, 1996. Proceedings CVPR '96, 1996 IEEE Computer
  Society Conference on}, Jun 1996, pp. 183--188.

\bibitem{capurro1997}
C.~Capurro, F.~Panerai, and G.~Sandini, ``Dynamic vergence using log-polar
  images,'' \emph{International Journal of Computer Vision}, vol.~24.

\bibitem{panerai98}
F.~Panerai and G.~Sandini, ``Oculo-motor stabilization reflexes: integration of
  inertial and visual information,'' \emph{Neural Networks}, vol.~11, no. 7-8,
  pp. 1191--1204, 1998.

\bibitem{shibata2000}
T.~Shibata and S.~Schaal, \emph{Biomimetic gaze stabilization}.\hskip 1em plus
  0.5em minus 0.4em\relax World Scientific, 2000, pp. 31--52.

\bibitem{panerai2002}
F.~Panerai, G.~Metta, and G.~Sandini, ``Learning visual stabilization reflexes
  in robots with moving eyes,'' \emph{Neurocomputing}, vol.~48, no. 1–4, pp.
  323 -- 337, 2002.

\bibitem{gay2012}
S.~Gay, A.~Ijspeert, and J.~Santos-Victor, ``Predictive gaze stabilization
  during periodic locomotion based on adaptive frequency oscillators,'' in
  \emph{Robotics and Automation (ICRA), 2012 IEEE International Conference on},
  May 2012, pp. 271--278.

\bibitem{oliveira2009}
C.~P. Santos, M.~Oliveira, A.~M.~A. Rocha, and L.~Costa, ``Head motion
  stabilization during quadruped robot locomotion: Combining dynamical systems
  and a genetic algorithm,'' in \emph{Robotics and Automation, 2009. ICRA '09.
  IEEE International Conference on}, May 2009, pp. 2294--2299.

\bibitem{UgoPhdThesis}
U.~Pattacini, ``Modular cartesian controllers for humanoid robots: Design and
  implementation on the i{C}ub,'' Ph.D. dissertation, Istituto Italiano di
  Tecnologia, Genova, Italy, 2011.

\bibitem{MettaiCub2010}
G.~Metta, L.~Natale, F.~Nori, G.~Sandini, D.~Vernon, L.~Fadiga, C.~von Hofsten,
  K.~Rosander, M.~Lopes, J.~Santos-Victor, A.~Bernardino, and L.~Montesano,
  ``The {iCub} humanoid robot: An open-systems platform for research in
  cognitive development,'' \emph{Neural Networks}, vol.~23, no. 8-9, pp.
  1125--1134, 2010.

\bibitem{xsense}
\BIBentryALTinterwordspacing
{X}sens website. [Online]. Available: \url{http://www.xsens.com/}
\BIBentrySTDinterwordspacing

\bibitem{opencv_library}
G.~Bradski, ``{The OpenCV Library},'' \emph{Dr. Dobb's Journal of Software
  Tools}, 2000.

\bibitem{farneback2003}
G.~Farneb\"{a}ck, ``\BIBforeignlanguage{English}{Two-frame motion estimation
  based on polynomial expansion},'' in \emph{\BIBforeignlanguage{English}{Image
  Analysis}}, ser. Lecture Notes in Computer Science, J.~Bigun and
  T.~Gustavsson, Eds.\hskip 1em plus 0.5em minus 0.4em\relax Springer Berlin
  Heidelberg, 2003, vol. 2749, pp. 363--370.

\end{thebibliography}

\end{document}